%% file: main.tex
\pdfoutput=1
\documentclass[11pt]{article}

\usepackage[]{acl}

\usepackage{times}
\usepackage{latexsym}

\usepackage[T1]{fontenc}
\usepackage[utf8]{inputenc}

\usepackage{microtype}

\usepackage{subcaption}
\usepackage{graphicx}
\usepackage{booktabs}
\usepackage{multicol}
\usepackage{multirow}
\usepackage{amsmath}
\usepackage{amssymb}

\input{settings.tex}

\input{math_commands.tex}

\title{Knowledge Neurons in Pretrained Transformers}

\author{Damai Dai$^{\dag\ddag}$\thanks{\ \  Contribution during internship at Microsoft Research.},~~Li Dong$^\ddag$,~~Yaru Hao$^\ddag$,~~Zhifang Sui$^\dag$,~~Baobao Chang$^\dag$,~~Furu Wei$^\ddag$\\
$^\dag$MOE Key Lab of Computational Linguistics, Peking University \\
$^\ddag$Microsoft Research \\
\texttt{\{daidamai,szf,chbb\}@pku.edu.cn}
\\\texttt{\{lidong1,yaruhao,fuwei\}@microsoft.com} \\
}

\begin{document}
\maketitle
\begin{abstract}
Large-scale pretrained language models are surprisingly good at recalling factual knowledge presented in the training corpus~\citep{lama,lpaqa}. 
In this paper, we present preliminary studies on how factual knowledge is stored in pretrained Transformers by introducing the concept of \textit{knowledge neurons}. 
Specifically, we examine the fill-in-the-blank cloze task for BERT. Given a relational fact, we propose a knowledge attribution method to identify the neurons that express the fact. 
We find that the activation of such knowledge neurons is positively correlated to the expression of their corresponding facts. 
In our case studies, we attempt to leverage knowledge neurons to edit (such as update, and erase) specific factual knowledge without fine-tuning. 
Our results shed light on understanding the storage of knowledge within pretrained Transformers. 
The code is available at \url{https://github.com/Hunter-DDM/knowledge-neurons}. 
\end{abstract}

\section{Introduction}
\label{sec:intro}

Large-scale pretrained Transformers~\citep{bert,roberta,unilm,electra,unilmv2} are usually learned with a language modeling objective on large-scale corpora, such as Wikipedia, where exists oceans of factual knowledge.
Pretrained language models naturally play as a free-text knowledge base by predicting texts~\citep{comet}.
\citet{lama} and \citet{lpaqa} probe factual knowledge stored in pretrained language models by fill-in-the-blank cloze queries. 
The evaluation shows that pretrained Transformers have a strong ability to recall factual knowledge without any fine-tuning.
\citet{knowledge_pack} use closed-book question answering to show that the larger a model is, the more knowledge it can store. 
However, most previous work focuses on evaluating the overall accuracy of text-form knowledge prediction.
In this paper, we attempt to look deeper into pretrained Transformers and investigate how factual knowledge is stored.

\begin{figure}[t]
\centering
\includegraphics[width=0.9\linewidth]{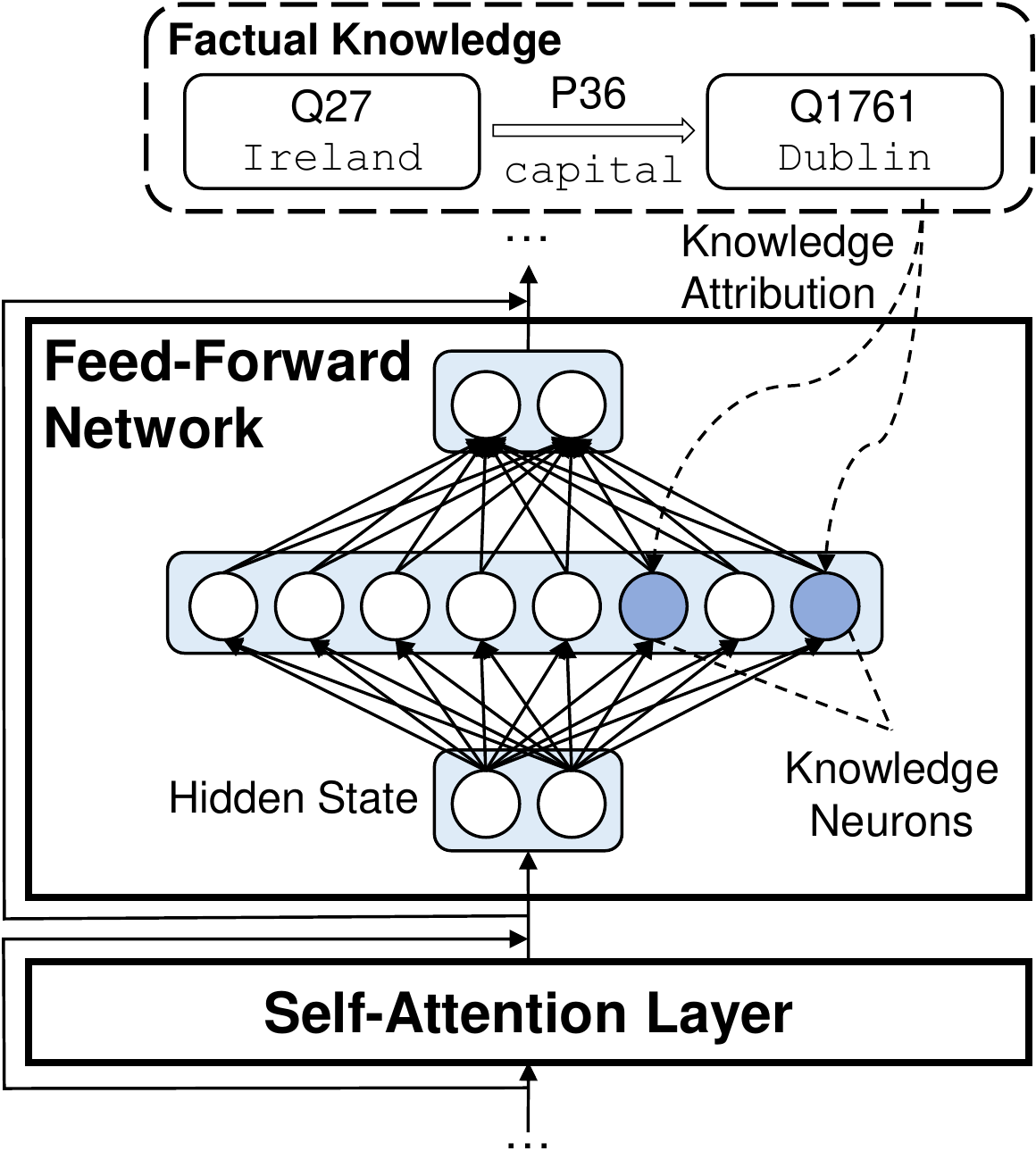}
\caption{
Through knowledge attribution, we identify knowledge neurons that express a relational fact. 
}
\label{fig:know_attr}
\end{figure}

As shown in Figure~\ref{fig:know_attr}, we propose a knowledge attribution method to identify the neurons that express a relational fact, where such neurons are named \textit{knowledge neurons}. 
Specifically, we view feed-forward network (i.e., two-layer perceptron) modules in Transformer as key-value memories~\citep{ffn_memory}. 
For the example in Figure~\ref{fig:know_attr}, the hidden state is fed into the first linear layer and activates knowledge neurons; then, the second linear layer integrates the corresponding memory vectors.
The key-value-memory nature~\citep{ffn_memory} inspires us to propose the knowledge attribution method, which identifies knowledge neurons in feed-forward networks by computing the contribution of each neuron to the knowledge prediction. 

\begin{figure*}[t]
\centering
\includegraphics[width=0.99\linewidth]{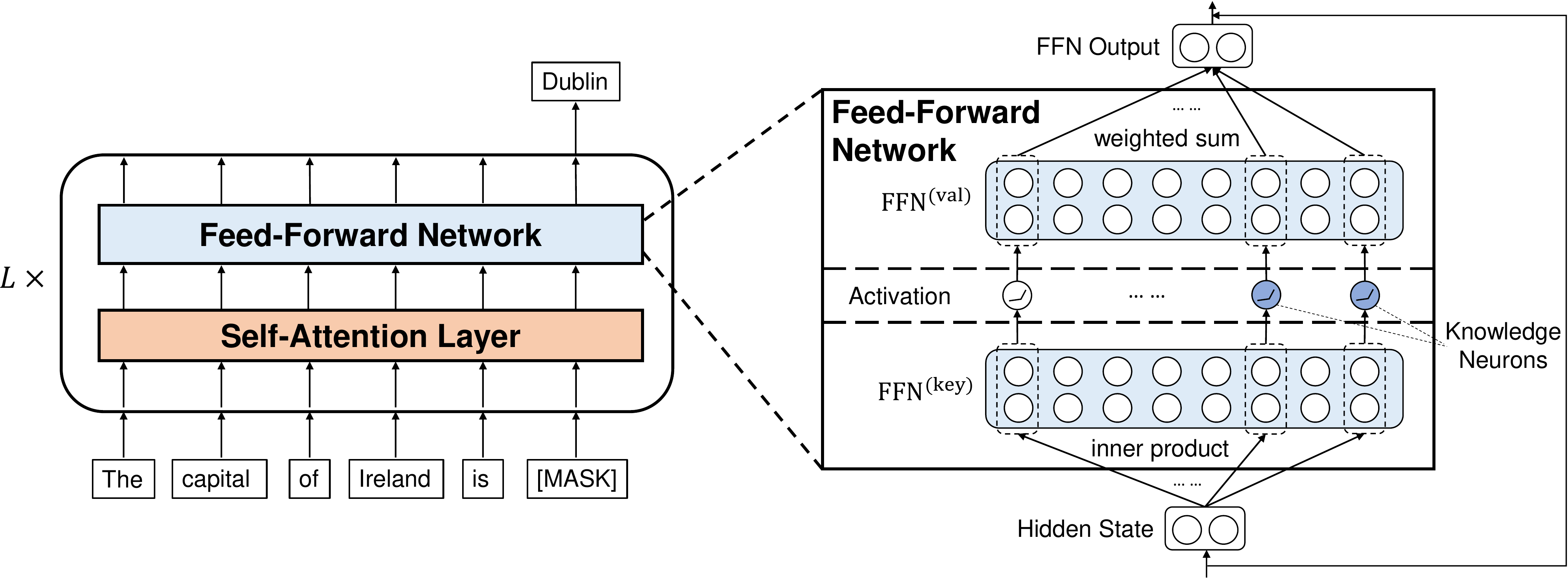}
\caption{
Illustration of how an FFN module in a Transformer block works as a key-value memory. 
The first linear layer $\operatorname{FFN^{(key)}}$ computes intermediate neurons through inner product. 
Taking the activation of these neurons as weights, the second linear layer $\operatorname{FFN^{(val)}}$ integrates value vectors through weighted sum. 
We hypothesize that knowledge neurons in the FFN module are responsible for expressing factual knowledge. 
}
\label{fig:ffn_as_memory}
\end{figure*}

Extensive analysis shows that the activation of the identified knowledge neurons is positively correlated to the knowledge expression, which shows the effectiveness of the proposed knowledge attribution method.
First, suppressing and amplifying knowledge neurons notably affects the expression of the corresponding knowledge. 
Second, we find that knowledge neurons of a fact tend to be activated more by corresponding knowledge-expressing prompts. 
Third, given the knowledge neurons of a fact, the top activating prompts retrieved from open-domain texts usually express the corresponding fact, while the bottom activating prompts do not express the correct relation.

In our case studies, we try to leverage knowledge neurons to explicitly edit factual knowledge in pretrained Transformers without any fine-tuning.
We present two preliminary studies: updating facts, and erasing relations.
After identifying the knowledge neurons, we perform a knowledge surgery for pretrained Transformers by directly modifying the corresponding parameters in feed-forward networks.
% successfully updates and erases the target knowledge
Such surgery shows promising results, keeping a moderate influence on other knowledge.

Our contributions are summarized as follows: 
\begin{itemize}
\item We introduce the concept of \textit{knowledge neurons} and propose a knowledge attribution method to identify the knowledge neurons that express specific factual knowledge in the fill-in-the-blank cloze task. 
\item We conduct both qualitative and quantitative analysis to show that knowledge neurons are positively correlated to knowledge expression.
\item We present preliminary studies of leveraging knowledge neurons to edit factual knowledge in Transformers, even without any fine-tuning.
\end{itemize}

\section{Background: Transformer}
\label{sec:background}

Transformer~\citep{transformer} is one of the most popular and effective NLP architectures. 
A Transformer encoder is stacked with $L$ identical blocks. 
Each Transformer block mainly contains two modules: a self-attention module, and a feed-forward network~(abbreviated as FFN) module. 
Let $X \in \mathbb{R}^{n \times d}$ denote the input matrix, two modules can be formulated as follows: 
\begin{align}
Q_{h} = X W_{h}^{Q}, & K_{h} = X W_{h}^{K}, V_{h} = X W_{h}^{V}, \\
\operatorname{Self-Att}_{h}(X) & = \operatorname{softmax} \left(Q_{h} K_{h}^{T} \right) V_{h}, \label{equ:self_att} \\
\operatorname{FFN}(H) & = \operatorname{gelu} \left(H W_1 \right) W_2, \label{equ:ffn}
\end{align}
where $W_{h}^{Q}, W_{h}^{K}, W_{h}^{V}, W_1, W_2$ are parameter matrices; 
$\operatorname{Self-Att}_{h}(X)$ computes a single attention head; 
$H$, the hidden state, is given by projecting the concatenation of all heads; 
$\operatorname{gelu}$ denotes the GELU activation function~\citep{gelu}.
For simplicity, we omit the scaling factor in self-attention and the bias terms. 

\paragraph{Connections Between Self-Attention and FFN}
Comparing \eqform{equ:self_att} and \eqform{equ:ffn}, we notice that the formula of $\operatorname{FFN}(\cdot)$ is quite similar to $\operatorname{Self-Att}(\cdot)$, except the activation function $\operatorname{gelu}$ in FFN and $\softmax$ in self-attention.
Thus, similar to the query-key-value mechanism in self-attention, it is reasonable to regard the input of the FFN as a query vector, and two linear layers of the FFN as keys and values, respectively.
Similar observations are also described in~\citep{ffn_memory}.

\section{Identifying Knowledge Neurons}
\label{sec:identify_knowledge_neurons}

Similar to \citep{ffn_memory}, we view FFNs in Transformer as key-value memories as illustrated in Figure~\ref{fig:ffn_as_memory}. 
We hypothesize that factual knowledge is stored in FFN memories and expressed by \textit{knowledge neurons}.
In this section, we propose a knowledge attribution method and a refining strategy to identify these knowledge neurons. 

\subsection{Knowledge Assessing Task}

We employ the fill-in-the-blank cloze task to assess whether a pretrained model knows a fact.
Following~\citet{lama}, each relational fact is in the form of a triplet $\langle h, r, t \rangle$, where $h$ is the head entity, $t$ is the tail entity, and $r$ is the relation between them. 
Given a fact, pretrained models answer the cloze query $x$ that expresses the fact but leaves the tail entity as a blank. 
For example, given the fact $\langle \texttt{Ireland, capital, Dublin} \rangle$, a possible query is ``\textit{The capital of Ireland is \underline{~~~~}}''. 
We also call the query a \textit{knowledge-expressing prompt}. 
\citet{lama} describe that a model knows a fact if it can predict the correct answer. 
In this paper, rather than just examining the model outputs, we identify the specific knowledge neurons that express factual knowledge.

\subsection{Knowledge Attribution}

Inspired by~\citet{attattr}, we propose a knowledge attribution method based on integrated gradients~\citep{ig}. 
Our method can evaluate the contribution of each neuron to knowledge predictions. 
In this paper, we examine FFN intermediate neurons for the masked token, where the answer is predicted. 

Given an input prompt $x$, we first define the model output $\operatorname{P}_x(\hat{w}^{(l)}_{i})$ as the probability of the correct answer predicted by a pretrained model: 
\begin{equation}
    \operatorname{P}_x(\hat{w}^{(l)}_{i}) = p(y^* | x, w^{(l)}_{i}=\hat{w}^{(l)}_{i}), 
\end{equation}
where $y^*$ denotes the correct answer; 
$w^{(l)}_{i}$ denotes the $i$-th intermediate neuron in the $l$-th FFN; 
$\hat{w}^{(l)}_{i}$ is a given constant that $w^{(l)}_{i}$ is assigned to. 

In order to calculate the attribution score of a neuron $\operatorname{Attr}(w^{(l)}_{i})$, we gradually change $w^{(l)}_{i}$ from $0$ to its original value $\overline{w}^{(l)}_{i}$ calculated by the pretrained model, and meanwhile integrate the gradients: 
\begin{equation}
\operatorname{Attr}(w^{(l)}_{i}) = \overline{w}^{(l)}_{i} \int_{\alpha=0}^{1} \frac{\partial \operatorname{P}_x(\alpha \overline{w}^{(l)}_{i})}{\partial w^{(l)}_{i}} d \alpha,
\end{equation}
where $\frac{\partial \operatorname{P}_x(\alpha \overline{w}^{(l)}_{i})}{\partial w^{(l)}_{i}}$ calculates the gradient of the model output with regard to $w^{(l)}_{i}$. 
Intuitively, as $\alpha$ changes from $0$ to $1$, by integrating the gradients, $\operatorname{Attr}(w^{(l)}_{i})$ accumulates the output probability change caused by the change of $w^{(l)}_{i}$.
If the neuron has a great influence on the expression of a fact, the gradient will be salient, which in turn has large integration values.
Therefore, the attribution score can measure the contribution of the neuron $w^{(l)}_{i}$ to the factual expressions.

Directly calculating continuous integrals is intractable. 
We instead use Riemann approximation $\Tilde{\operatorname{Attr}}(w^{(l)}_{i}) = \frac{\overline{w}^{(l)}_{i}}{m} \sum_{k=1}^{m} \frac{\partial \operatorname{P}_x(\frac{k}{m} \overline{w}^{(l)}_{i})}{\partial w^{(l)}_{i}}, $
where $m=20$ is the number of approximation steps.
% We empirically set $m$ to $20$.
With the attribution algorithm, we can identify a coarse set of knowledge neurons whose attribution scores are greater than a threshold $t$. 

\begin{table*}[t]
\footnotesize
\centering
\setlength{\tabcolsep}{1.5pt}
\begin{tabular}{l l l l}
\toprule
\textbf{Relations} & \textbf{Template \#1} &  \textbf{Template \#2} & \textbf{Template \#3} \\ 
\midrule
P176 (\texttt{manufacturer}) & [X] is produced by [Y] & [X] is a product of [Y] & [Y] and its product [X] \\
P463 (\texttt{member\_of}) & [X] is a member of [Y] & [X] belongs to the organization of [Y] & [X] is affiliated with [Y] \\
P407 (\texttt{language\_of\_work}) & [X] was written in [Y] & The language of [X] is [Y] & [X] was a [Y]-language work \\
\bottomrule
\end{tabular}
\caption{
Example prompt templates of three relations in \textsc{ParaRel}. 
[X] and [Y] are the placeholders for the head and tail entities, respectively. 
Owing to the page width, we show only three templates for each relation. 
Prompt templates in \textsc{ParaRel} produce 253,448 knowledge-expressing prompts in total for 27,738 relational facts. 
}
\label{tab:prompt_example}
\end{table*}

\subsection{Knowledge Neuron Refining}
\label{sec:knowledge_refining}

In order to identify knowledge neurons more accurately, we further propose a refining strategy. 
Besides ``true-positive'' knowledge neurons that express factual knowledge, the coarse set of knowledge neurons may contain ``false-positive'' knowledge neurons that express other information (e.g., syntactic or lexical information). 
The refining strategy aims to filter out these ``false-positive'' neurons.

For different prompts corresponding to the same fact, we hypothesize that they share the same set of ``true-positive'' knowledge neurons, since they express the same factual knowledge.
Meanwhile, we hypothesize that they do not share the ``false-positive'' knowledge neurons as long as the prompts are diverse enough.
Therefore, given multiple diverse prompts, we can refine the coarse set of knowledge neurons by retaining only neurons that are widely shared among these prompts.

Specifically, given a relational fact, the complete process to identify its knowledge neurons is described as follows: 
(1) produce $n$ diverse prompts;
(2) for each prompt, calculate the knowledge attribution scores of neurons;
(3) for each prompt, retain the neurons with attribution scores greater than the attribution threshold $t$, obtaining the coarse set of knowledge neurons;
(4) considering all the coarse sets together, retain the knowledge neurons shared by more than $p\%$ prompts.

\section{Experiments}

\subsection{Experimental Settings}

We conduct experiments for BERT-base-cased~\citep{bert}, one of the most widely-used pretrained models. 
It contains 12 Transformer blocks, where the hidden size is 768 and the FFN inner hidden size is 3,072. 
Notice that our method is not limited to BERT and can be easily generalized to other pretrained models. 
For each prompt, we set the attribution threshold $t$ to $0.2$ times the maximum attribution score. 
For each relation, we initialize the refining threshold $p\%$ (Section~\ref{sec:knowledge_refining}) as $0.7$.
Then, we increase or decrease it by $0.05$ at a time until the average number of knowledge neurons lies in [2, 5].
We run our experiments on NVIDIA Tesla V100 GPUs. 
On average, it costs 13.3 seconds to identify knowledge neurons for a relational fact with 9 prompts.

\subsection{Dataset}

We examine knowledge neurons through the fill-in-the-blank cloze task based on the \textsc{ParaRel} dataset~\citep{consistency}. 
\textsc{ParaRel} is curated by experts, containing various prompt templates for 38 relations from the T-REx dataset~\citep{trex}. 
We show some example templates in Table~\ref{tab:prompt_example}. 
For each relational fact, we fill in the head entity in prompt templates and leave the tail entity as a blank to predict.
In order to guarantee the template diversity, we filter out relations with fewer than 4 prompt templates and finally keep 34 relations, where each relation has 8.63 different prompt templates on average. 
These prompt templates produce 253,448 knowledge-expressing prompts in total for 27,738 relational facts. 

\subsection{Attribution Baseline}
\label{sec:baseline}

Our baseline method takes the neuron activation value as the attribution score, i.e., $\operatorname{Attr}_{\text{base}}(w^{(l)}_{i}) = \overline{w}^{(l)}_{i}$, which measures how sensitive a neuron is to the input.
After computing attribution scores, we follow the same pipeline to obtain the refined knowledge neurons.
For a fair comparison, we employ the same method to choose the hyper-parameters $t$ and $p\%$ for the baseline to ensure the average number of knowledge neurons for each relation lies in $[2, 5]$.

The method based on neuron activation is a reasonable baseline.
It is motivated by FFNs's analogy with the self-attention mechanism (as described in Section~\ref{sec:background}), because self-attention scores are usually used as a strong attribution baseline~\citep{kovaleva-etal-2019-revealing,analysis_trm_att_1,attattr}.

\subsection{Statistics of Knowledge Neurons}
\label{sec:distribution}

\begin{figure}[t]
\centering
\includegraphics[width=0.98\linewidth]{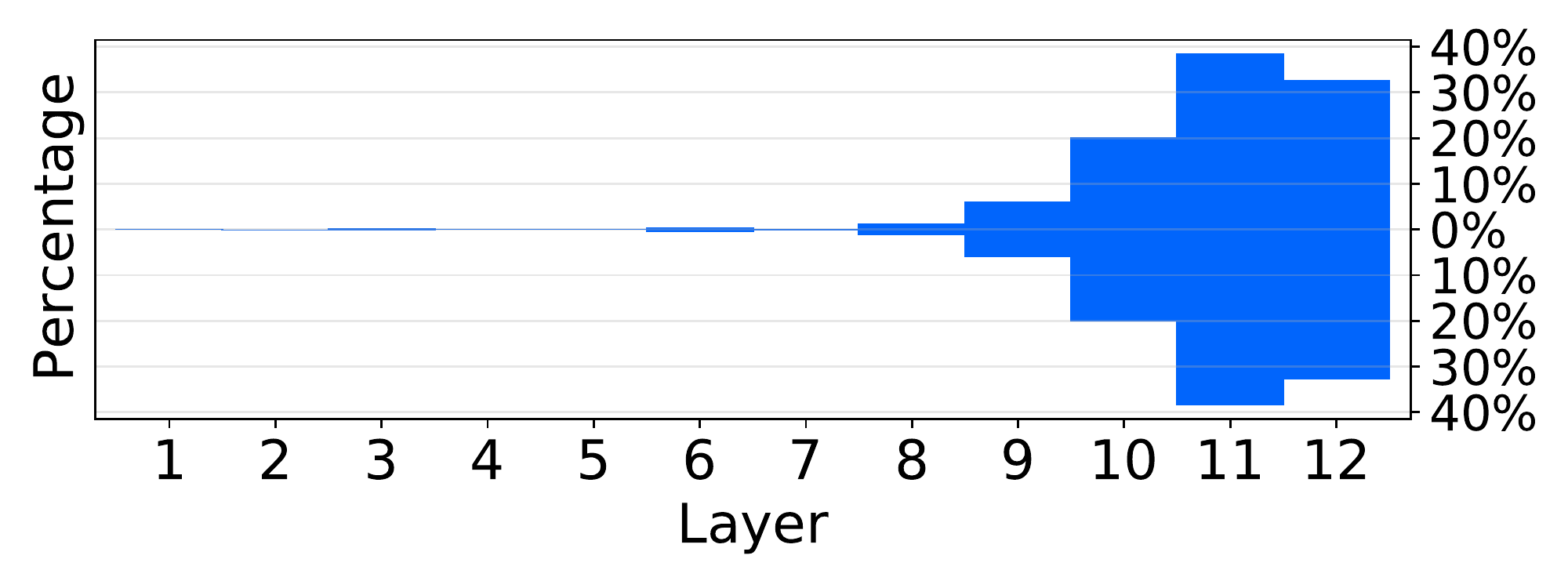}
\caption{
Percentage of knowledge neurons identified by our method in each Transformer layer. 
}
\label{fig:distribution}
\end{figure}

Figure~\ref{fig:distribution} presents the layer distribution of knowledge neurons identified by our knowledge attribution method.
We notice that most fact-related neurons are distributed in the topmost layers of pretrained Transformers.
The finding also agrees with \citet{bert:rediscover} and \citet{ffn_memory}.

\begin{table}[t]
% \footnotesize
\centering
\setlength{\tabcolsep}{7pt}
\begin{tabular}{l l l}
\toprule
\textbf{Type of Neurons}& \textbf{Ours} & \textbf{Baseline} \\ 
\midrule
Knowledge neurons & 4.13 & 3.96 \\ 
$\bigcap$ of intra-rel. fact pairs & 1.23 & 2.85 \\
$\bigcap$ of inter-rel. fact pairs & 0.09 & 1.92 \\ 
\bottomrule
\end{tabular}
\caption{
Statistics of knowledge neurons. 
$\bigcap$ denotes the intersection of knowledge neurons of fact pairs. 
``\textit{rel.}'' is the shorthand of relation. 
Our method identifies more exclusive knowledge neurons.
}
\label{tab:distribution}
\end{table}

\begin{figure*}[t]
\centering
\includegraphics[width=0.99\linewidth]{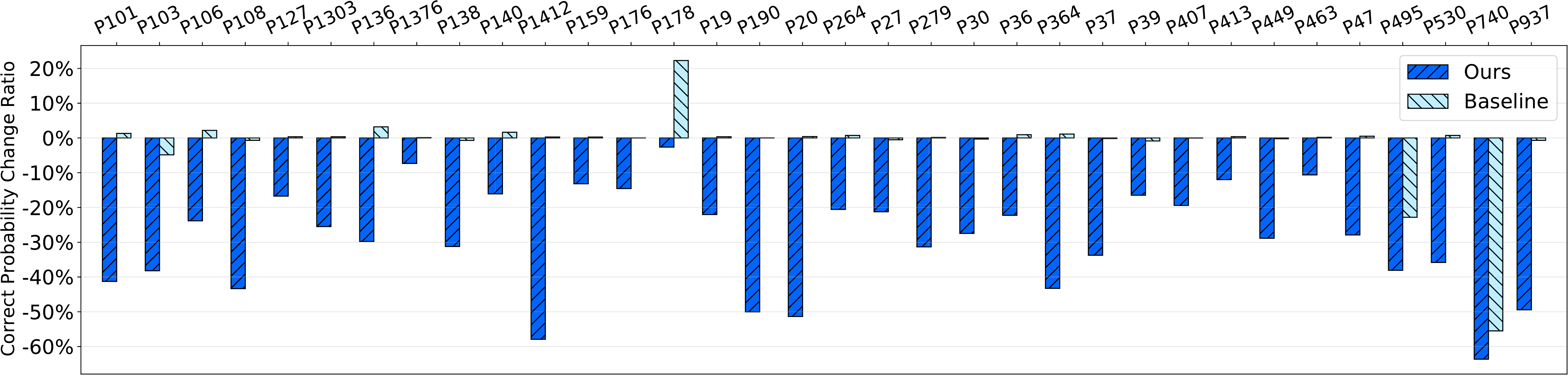}
\caption{
Results of suppressing knowledge neurons for various relations. 
Suppressing knowledge neurons decreases the correct probability by 29.03\% on average. 
For the baseline, the decreasing ratio is 1.47\% on average. 
}
\label{fig:suppress}
\end{figure*}

\begin{figure*}[t]
\centering
\includegraphics[width=0.99\linewidth]{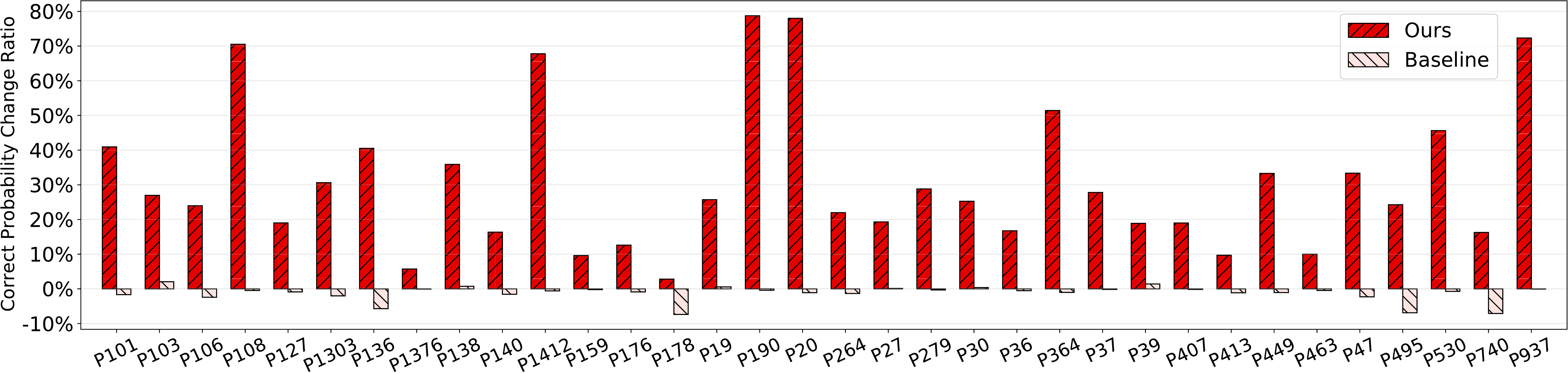}
\caption{
Results of amplifying knowledge neurons for various relations.
Amplifying knowledge neurons increases the correct probability by 31.17\% on average. 
For the baseline, the correct probability even decreases by 1.27\%.
}
\label{fig:amplify}
\end{figure*}

Table~\ref{tab:distribution} shows statistics of knowledge neurons.
On average, we identify $4.13$ knowledge neurons for each relational fact using our knowledge attribution method, and $3.96$ using the baseline method.
Their same order of magnitude guarantees the fairness of the subsequent comparisons in the paper.

We also compute the knowledge neuron intersection of different relational facts. 
Table~\ref{tab:distribution} shows the average number of pair-wise knowledge neuron intersections. 
For our proposed method, (1) fact pairs with the same relation~(\textit{intra-relation fact pairs}) share 1.23 knowledge neurons on average; (2) fact pairs with different relations~(\textit{inter-relation fact pairs}) share almost no knowledge neurons. 
In contrast, for the baseline, (3) most identified neurons are shared by intra-relation fact pairs; (4) even a substantial portion of neurons are common for inter-relation fact pairs. 
The difference in knowledge neuron intersections suggests that our method can identify more exclusive knowledge neurons.

\begin{table*}[t]
\footnotesize
\centering
\setlength{\tabcolsep}{3.5pt}
\begin{tabular}{l c l l}
\toprule
\textbf{Relational Facts} & \textbf{Neurons} & \multicolumn{2}{c}{\textbf{Top-2 and Bottom-2 Activating Prompts (Average Activation)}} \\
\midrule
% P36-229
\multirow{5}{*}{\tabincell{l}{$\langle$ \texttt{Ireland}, \\ ~~ \texttt{capital}, \\ ~~~~\texttt{Dublin} $\rangle$}} & \multirow{5}{*}{\tabincell{l}{$w^{(9)}_{2141},w^{(10)}_{1122}$}} & \multirow{2}{*}{Top} & Our trip ... in \textbf{\underline{Dublin}}, the capital and largest city of \textbf{Ireland} ... (6.36) \\
 & & & \textbf{\underline{Dublin}} is the capital and largest city of \textbf{Ireland}. (5.77) \\
\cmidrule{3-4}
 & & \multirow{2}{*}{Bottom} & \textbf{\underline{Dublin}} just might be the most iconic destination in all of \textbf{Ireland}. (1.27) \\
 & & & ... in \textbf{Ireland}'s famed city, you can enjoy ... \textbf{\underline{Dublin}} experience ... (-0.30) \\
\midrule
% P19-129
\multirow{5}{*}{\tabincell{l}{$\langle$ \texttt{Cao\_Yunding}, \\ ~~ \texttt{place\_of\_birth}, \\ ~~~~\texttt{Shanghai} $\rangle$}} & \multirow{5}{*}{\tabincell{l}{$w^{(10)}_{739},w^{(10)}_{1885},$\\$w^{(11)}_{2876}$}} & \multirow{2}{*}{Top} & \textbf{Cao Yunding} was born in \textbf{\underline{Shanghai}} in November 1989. (3.58) \\
 & & & Full name: \textbf{Cao Yunding} ... Place of birth: \textbf{\underline{Shanghai}}, China ... (2.73) \\
\cmidrule{3-4}
 & & \multirow{2}{*}{Bottom} & ... \textbf{Cao Yunding} (\textbf{\underline{Shanghai}} Shenhua) is shown the red card ... (-0.30) \\
 & & & \textbf{\underline{Shanghai}} Shenhua midfielder \textbf{Cao Yunding} ... (-0.31) \\
\midrule
% P30-557
\multirow{5}{*}{\tabincell{l}{$\langle$ \texttt{Kuwait}, \\ ~~ \texttt{continent}, \\ ~~~~\texttt{Asia} $\rangle$}} & \multirow{5}{*}{\tabincell{l}{$w^{(6)}_{147},w^{(9)}_{866},$ \\ $w^{(9)}_{1461},w^{(10)}_{1169}$}} & \multirow{2}{*}{Top} & \textbf{Kuwait} is thus one of the smallest countries in \textbf{\underline{Asia}} ... (6.63) \\
 & & & \textbf{Kuwait} is a country in Western \textbf{\underline{Asia}} ... (6.27) \\
\cmidrule{3-4}
 & & \multirow{2}{*}{Bottom} & This page displays all \textbf{\underline{Asia}} Society content on \textbf{Kuwait} ... (-0.48) \\
 & & & Noor \textbf{\underline{Asia}} is ... distribution companies in \textbf{Kuwait} ... (-0.59) \\
\bottomrule
\end{tabular}
\caption{
Example relational facts along with their knowledge neurons, their top-2 and bottom-2 activating prompts, and the corresponding neuron activation. 
$w^{(l)}_{i}$ denotes the $i$-th intermediate neuron at the $l$-th FFN. 
We fill the blank in each prompt with the correct answer for better readability. 
Owing to the page width, we show only key parts for overlong prompts. 
The top-2 activating prompts express exactly the relation, but the bottom-2 do not. 
}
\label{tab:trigger_example}
\end{table*}

\subsection{Knowledge Neurons Affect Knowledge Expression}
\label{sec:control}

We investigate how much knowledge neurons can affect knowledge expression in Figure~\ref{fig:suppress} and Figure~\ref{fig:amplify}. 
Given a relational fact, we manipulate its knowledge neurons in two ways: 
(1) suppressing knowledge neurons by setting their activations to 0;
(2) amplifying knowledge neurons by doubling their activations. 
Then, for each relation, we plot the average change ratio of the probability for the correct answer, corresponding to the manipulation. 
For comparison, we also plot the results of manipulating baseline-identified knowledge neurons. 

Figure~\ref{fig:suppress} shows that suppressing knowledge neurons identified by our knowledge attribution method leads to a consistent decrease (29.03\% on average) in the correct probability. 
By contrast, for baseline-identified neurons, the suppressing operation has a negligible influence (1.47\% decrease on average) on the correct probability.
Notably, for the relation P178 (\texttt{developer}), the correct probability abnormally increases by using the baseline.

As shown in Figure~\ref{fig:amplify}, we have similar observations for amplifying the knowledge neurons identified by our knowledge attribution.
We see a consistent increase (31.17\% on average) in the correct probability. 
By contrast, the baseline even decreases the average correct probability by 1.27\%.

In summary, the knowledge neurons identified by our knowledge attribution method tend to notably affect knowledge expression.
Notice that the above assessment is affected by the distribution of knowledge neurons.
For example, if the knowledge neurons for a relation are distributed more widely, we need to manipulate more top-$k$ neurons for better control.
We use the above experiments as a proof of concept while leaving precise control for future work.

\begin{table}[t]
% \footnotesize
\centering
\setlength{\tabcolsep}{3pt}
\begin{tabular}{l c c}
\toprule
\textbf{Prompt Types} & \textbf{Ours} & \textbf{Baseline} \\ 
\midrule
Containing head and tail ($\mathcal{T}_{1}$) & 0.485 & 2.472 \\ 
Containing only head ($\mathcal{T}_{2}$) & 0.019 & 2.312 \\ 
Randomly sampled ($\mathcal{T}_{3}$) & -0.018 & 2.244 \\ 
\bottomrule
\end{tabular}
\caption{
Average activation of knowledge neurons for three types of prompts. 
The activation of neurons identified by our method can distinguish the knowledge-expressing prompts ($\mathcal{T}_{1}$) clearly. 
}
\label{tab:distant}
\end{table}

\subsection{Knowledge Neurons are Activated by Knowledge-Expressing Prompts}
\label{sec:activate}

In order to study what prompts can activate knowledge neurons, we compare the average activation of knowledge neurons for different types of prompts.

\paragraph{\textsc{BingRel} Dataset}
We build a new dataset \textsc{BingRel} by crawling the Bing search engine to collect new prompts, for a more extensive comparison beyond the \textsc{ParaRel} dataset.
For each of the 27,738 facts in \textsc{ParaRel}, we crawl two types of texts:
(1) up to ten texts containing both the head and the tail entities (210,217 texts crawled in total);
(2) up to ten texts containing only the head entity without restricting tail entities (266,020 texts crawled in total). 
Following the distant supervision assumption~\citep{distant_supervision}, the first type of texts tends to express the whole relational fact, while the second type does not.
We mask tail entities for the first type of texts to obtain knowledge-expressing prompts ($\mathcal{T}_{1}$).
In order to conduct a controlled experiment, we mask random words for the second type of texts, forming a control group ($\mathcal{T}_{2}$).
Moreover, we employ randomly sampled prompts as another control group ($\mathcal{T}_{3}$).

\paragraph{Results}
As shown in Table~\ref{tab:distant}, for our method, the identified knowledge neurons are more significantly activated by knowledge-expressing prompts ($\mathcal{T}_{1}=0.485$), compared with the control groups ($\mathcal{T}_{2}=0.019$ and $\mathcal{T}_{3}=-0.018$). 
By contrast, for the baseline, the activation of identified neurons cannot distinguish three types of prompts.  
In addition, since our comparison is based on the web-crawled \textsc{BingRel} dataset, we validate the generalization of knowledge neurons to open-domain texts that are unseen in \textsc{ParaRel}.  

\begin{table*}[t]
\footnotesize
\centering
\setlength{\tabcolsep}{7pt}
\begin{tabular}{l c c c c}
\toprule
\multirow{2}{*}{\textbf{Erased Relations}} & \multicolumn{2}{c}{\textbf{Perplexity (Erased Relation)}} & \multicolumn{2}{c}{\textbf{Perplexity (Other Relations)}} \\ 
\cmidrule(lr){2-3}\cmidrule(lr){4-5}
 & \textbf{Before Erasing} & \textbf{After Erasing} & \textbf{Before Erasing} & \textbf{After Erasing} \\
\midrule
P19 (\texttt{place\_of\_birth}) & 1450.0 & 2996.0 (+106.6\%) & 120.3 & 121.6 (+1.1\%) \\
P27 (\texttt{country\_of\_citizenship}) & ~~~~28.0 & ~~38.3 (+36.7\%) & 143.6 & 149.5 (+4.2\%) \\
P106 (\texttt{occupation}) & 2279.0 & 5202.0 (+128.2\%) & 120.1 & 125.3 (+4.3\%) \\
P937 (\texttt{work\_location}) & ~~~~58.0 & ~~140.0 (+141.2\%) & 138.0 & ~~151.9 (+10.1\%) \\
\bottomrule
\end{tabular}
\caption{
Case studies of erasing relations. 
The influence on knowledge expression is measured by the perplexity change. 
The knowledge erasing operation significantly affects the erased relation, and has just a moderate influence on the expression of other knowledge. 
}
\label{tab:erase_knowledge}
\end{table*}

\begin{table}[t]
\centering
\footnotesize
\setlength{\tabcolsep}{0.9pt}
\begin{tabular}{l c c}
\toprule
\textbf{Metric} & \textbf{Knowledge Neurons} & \textbf{Random Neurons} \\
\midrule
Change rate$\uparrow$ & 48.5\% & 4.7\% \\ 
Success rate$\uparrow$ & 34.4\% & 0.0\% \\ 
\midrule
% $\Delta$Intra-rel. PPL$\downarrow$ & 7.0\% & 8.5\% \\ 
% $\Delta$Inter-rel. PPL$\downarrow$ & 5.0\% & 3.2\% \\ 
% $\Delta$Intra-rel. PPL$\downarrow$ & 8.4 (120.6$\rightarrow$129.0) & 10.1 (118.8$\rightarrow$128.9) \\ 
% $\Delta$Inter-rel. PPL$\downarrow$ & 7.2 (144.5$\rightarrow$151.7) & ~~4.3 (134.3$\rightarrow$138.6) \\ 
$\Delta$Intra-rel. PPL$\downarrow$ & 8.4 & 10.1 \\ 
$\Delta$Inter-rel. PPL$\downarrow$ & 7.2 & 4.3 \\ 
\bottomrule
\end{tabular}
\caption{
Case studies of updating facts. 
$\uparrow$ means the higher the better, and $\downarrow$ means the lower the better.  
``\textit{rel.}'' is the shorthand of relation. 
Keeping a moderate influence on other knowledge, the surgery of knowledge neurons achieves a nontrivial success rate. 
}
\label{tab:update_knowledge}
\end{table}

\paragraph{Example Prompts}
In Table~\ref{tab:trigger_example}, we present example prompts that activate knowledge neurons the most and the least, respectively. 
Given a fact, we first identify its knowledge neurons with our knowledge attribution method. 
Then, we calculate the average activation of knowledge neurons for each crawled prompt that contains both the head and the tail entities in \textsc{BingRel}. 
Finally, we demonstrate two prompts with the highest average activation values and two with the lowest (denoted as top-2 and bottom-2 activating prompts, respectively). 
 
As shown in Table~\ref{tab:trigger_example}, the top-2 activating prompts express exactly the corresponding relational fact. 
In contrast, despite containing the same head and tail entities, the bottom-2 activating prompts do not express the correct relation. 
For example, although the bottom-2 activating prompts for $\langle \texttt{Ireland, capital, Dublin} \rangle$ express information like ``Dublin is a city in Ireland'', they do not reflect the \texttt{capital} relation. 
The examples support again that knowledge neurons are activated by corresponding knowledge-expressing prompts.

\section{Case Studies}
\label{sec:case:study}

We present two preliminary studies to demonstrate the potential applications of knowledge neurons.
We use the case studies as a proof of concept while leaving precise fact editing for future work.

\subsection{Updating Facts}
\label{sec:update_knowledge}

By leveraging knowledge neurons in pretrained models, we try to update a learned relational fact from $\langle h, r, t \rangle$ to $\langle h, r, t^{\prime} \rangle$.

\paragraph{Methods}
First, we identify the knowledge neurons of $\langle h, r, t \rangle$. 
Then, we retain the knowledge neurons that are shared by less than 10\% of intra-relation facts, to reduce the influence on other facts with the same relation. 
Finally, we directly modify the corresponding value slots in $\operatorname{FFN^{(val)}}$ (i.e., the second linear layer of FFNs; see Figure~\ref{fig:ffn_as_memory}): $\operatorname{FFN^{(val)}_i} = \operatorname{FFN^{(val)}_i} - \lambda_{1}\mathbf{t} + \lambda_{2}\mathbf{t^{\prime}}$, 
where $\operatorname{FFN^{(val)}_i}$ denotes the value slot corresponding to the $i$-th knowledge neuron; $\mathbf{t}$ and $\mathbf{t^{\prime}}$ are the word embeddings of $t$ and $t^{\prime}$, respectively; $\lambda_{1}$ and $\lambda_{2}$ are set to $1$ and $8$ in our experiments. 

\paragraph{Setup}
We conduct experiments on \textsc{ParaRel}.
For each relation, we randomly sample ten facts learned by the pretrained model.
For each fact $\langle h, r, t \rangle$, we randomly choose a different entity $t^{\prime}$ with the same type as $t$ (e.g., both $t$ and $t^{\prime}$ belong to \texttt{city}), and then update $t^{\prime}$ as the target entity.
We only manipulate about four top knowledge neurons as in Section~\ref{sec:distribution}.
For reference purposes, we also perform the same update process on the same number of random neurons.

\paragraph{Evaluation Metrics}
We report two metrics to evaluate the fact updating:
(1) {change rate}, the ratio that the original prediction $t$ is modified to another;
(2) {success rate}, the ratio that $t^{\prime}$ becomes the top prediction.
In addition, we measure the influence on other knowledge by the following two metrics:
(1) {$\Delta$intra-relation PPL}, the increase of perplexity on the prompts with the same relation $r$;
(2) {$\Delta$inter-relation PPL}, the increase of perplexity on the prompts with different relations.

\paragraph{Results}
As shown in Table~\ref{tab:update_knowledge}, the surgery of knowledge neurons achieves a nontrivial success rate for updating facts, while random neurons are insufficient.
Moreover, we find that such manipulation has little negative influence on other knowledge predictions.
It is promising that we can change very few (i.e., about four in the above experiments) neurons to affect certain facts in pretrained Transformers.
We can further improve the success rate by including more top knowledge neurons in the update process.

\subsection{Erasing Relations}
\label{sec:erase_knowledge}

We explore how to leverage knowledge neurons to erase specific relations in pretrained Transformers.
Specifically, we take four relations in \textsc{ParaRel} as examples, i.e., \texttt{place\_of\_birth}, \texttt{country\_of\_citizenship}, \texttt{occupation}, \texttt{work\_location}, that typically express sensitive personal information.

\paragraph{Methods}
Given a relation $r$, we first identify knowledge neurons for all relational facts with $r$.
Then, we retain $20$ knowledge neurons that appear most frequently among these facts.
Finally, we set the value slots in $\operatorname{FFN^{(val)}}$ (see Figure~\ref{fig:ffn_as_memory}) corresponding to these knowledge neurons to $\mathbf{0}$, i.e., zero vectors.

\paragraph{Results}
As shown in Table~\ref{tab:erase_knowledge}, we report model perplexity before and after knowledge erasing.
With the erasing operation, the perplexity of the removed knowledge increases as expected.
Moreover, the model perplexity of other relations remains similar.
We argue that knowledge neurons provide a promising way to erase undesired knowledge with minimal efforts.

\section{Related Work}

\paragraph{Probing Knowledge in Pretrained Models}

Many pieces of previous work aim to measure knowledge stored in pretrained models. 
\citet{lama} propose to retrieve knowledge in pretrained models (such as BERT) using cloze queries. 
Their experiments show that BERT has a strong ability to recall factual knowledge without any fine-tuning. 
\citet{lpaqa} improve the cloze queries with mining-based and paraphrasing-based methods. 
\citet{knowledge_pack} propose the closed-book question answering to measure how much knowledge a pretrained model has stored in its parameters.
\citet{consistency} measure and improve the consistency of pretrained models with respect to factual knowledge prediction.
Rather than examining only the model outputs, we provide an open-the-black-box analysis for the knowledge neurons in pretrained Transformers. 

\paragraph{Attribution Methods}

In order to open the black boxes of deep learning models, attribution methods aim to attribute the model output to input features using different measures. 
The product of the gradients (of the output with respect to input features) and feature values is a reasonable baseline~\citep{attr_baseline_1,attr_baseline_2}. 
Besides, a set of attribution methods~\citep{deeplift,lrp,deconvolutional_networks,guided_bp} back-propagate the final output to input features. 
However, as stated by~\citet{ig}, none of these methods can simultaneously satisfy \textit{sensitivity} and \textit{implementation invariance}, two fundamental axioms. 
Taking the axioms as guidance, \citet{ig} propose the integrated gradient method. 
Our knowledge attribution method is built upon integrated gradients.

\paragraph{Analysis of Transformer}

As one of the most popular and effective NLP architectures, Transformer~\citep{transformer} has attracted extensive studies. 
Most previous work focuses on the self-attention module~\citep{analysis_trm_att_1,analysis_trm_att_2,analysis_trm_att_3,attattr}. 
Recently, \citet{dynamic_conv} and~\citet{att_is_not_all_you_need} have pointed out that the feed-forward network module also matters to Transformer. 
\citet{ffn_memory} attempt to connect feed-forward networks with key-value memories by qualitative analysis. 
In this paper, we identify and analyze knowledge neurons in feed-forward networks for given factual knowledge. 
Moreover, we present how to leverage knowledge neurons to explicitly edit factual knowledge stored in pretrained Transformers.

\section{Conclusion and Future Directions}

We propose an attribution method to identify knowledge neurons that express factual knowledge in pretrained Transformers.
We find that suppressing or amplifying the activation of knowledge neurons can accordingly affect the strength of knowledge expression. 
Moreover, quantitative and qualitative analysis on open-domain texts shows that knowledge neurons tend to be activated by the corresponding knowledge-expressing prompts. 
In addition, we present two preliminary case studies that attempt to utilize knowledge neurons to update or erase knowledge in pretrained Transformers. 

Despite the effectiveness of identifying knowledge neurons, our current studies still have limitations. 
First, we examine knowledge neurons based on the fill-in-the-blank cloze task, while knowledge can be expressed in a more implicit way. 
It is an open question whether Transformer can utilize stored knowledge in a generalized way, such as for reasoning.
The interactions between knowledge neurons also remain under explored.
Second, we focus on factual knowledge for ease of evaluation, even though our method is also applicable for other types of knowledge. 
Third, we use the single-word blank in cloze queries for simplicity, which requires multi-word extensions~\citep{xfactr}. 
Besides, an interesting future direction is to figure out how knowledge neurons work in multilingual pretrained Transformers~\citep{xlm,xlmr,infoxlm}.

\section{Acknowledgement}

Damai Dai, Zhifang Sui, and Baobao Chang are supported by the National Key Research and Development Program of China 2020AAA0106701 and NSFC project U19A2065. 

\bibliographystyle{acl_natbib}
\bibliography{kneuron}

\end{document}

%% file: settings.tex
\usepackage{multirow}
\usepackage{amsmath}
\usepackage{capt-of}
\usepackage{tabularx}
\usepackage{epsfig}
\usepackage{amssymb}
\usepackage{amsfonts}
\usepackage{booktabs}
\usepackage{scalerel}
\usepackage[inline]{enumitem}
\usepackage{listings}
\usepackage{varwidth}
\usepackage[export]{adjustbox}
\usepackage{tikz}
\usetikzlibrary{tikzmark}
\usepackage{cleveref}

\usepackage{stmaryrd}
\usepackage{bbm}

\newcommand{\tabincell}[2]{\begin{tabular}{@{}#1@{}}#2\end{tabular}}

\newcommand{\eqform}[1]{Equation~(\ref{#1})}

\usepackage{algorithm}
\usepackage[noend]{algpseudocode}

\definecolor{deepblue}{rgb}{0,0,0.5}
\definecolor{officeblue}{RGB}{0,102,204}
\definecolor{deepred}{rgb}{0.6,0,0}
\definecolor{deepgreen}{rgb}{0,0.5,0}
\definecolor{mybrickred}{RGB}{182,50,28}

\definecolor{fillcolor}{RGB}{216,217,252}

\algnewcommand\algorithmicrequireb{{\hspace{0.85cm}}}
\algnewcommand\INPTDESCB{\item[\algorithmicrequireb]}

\algnewcommand\algorithmicfuncdesc{\textbf{Function:}}
\algnewcommand\FUNCDESC{\item[\algorithmicfuncdesc]}
\algnewcommand\algorithmicfuncdescb{{\hspace{1.48cm}}}
\algnewcommand\FUNCDESCB{\item[\algorithmicfuncdescb]}
\algnewcommand{\algorithmicgoto}{\textbf{goto}}
\algnewcommand{\Goto}[1]{\algorithmicgoto~\ref{#1}}

%% file: math_commands.tex
\usepackage{amsmath,amsfonts,bm}

\def\eqref#1{equation~\ref{#1}}
\def\1{\bm{1}}

\DeclareMathAlphabet{\mathsfit}{\encodingdefault}{\sfdefault}{m}{sl}
\SetMathAlphabet{\mathsfit}{bold}{\encodingdefault}{\sfdefault}{bx}{n}

\newcommand{\softmax}{\mathrm{softmax}}